\documentclass[lettersize,journal]{IEEEtran}
\usepackage{amsmath,amsfonts}
\usepackage{algorithmic}
\usepackage{algorithm}
\usepackage[algo2e]{algorithm2e}
\usepackage{array}
\usepackage{amssymb}
\usepackage[caption=false,font=footnotesize,labelfont=rm,textfont=rm]{subfig}
\usepackage{textcomp}
\usepackage{stfloats}
\usepackage{url}
\usepackage{verbatim}
\usepackage{graphicx}
\usepackage{xcolor}
\usepackage{color}
\usepackage{bm}
\usepackage{multirow}
\usepackage{hyperref}
\usepackage{subfig}
\usepackage{pifont}
\usepackage[square,numbers]{natbib}
\hypersetup{hypertex=true,colorlinks=true,anchorcolor=green,citecolor=green}
\begin{document}
\title{Rt-Track: Robust Tricks for Multi-Pedestrian Tracking}
\author{Yukuan~Zhang, Yunhua~Jia, Housheng~Xie, Mengzhen~Li
	
	Limin~Zhao, Shan~Zhao, Yang Yang,~\IEEEmembership{Member,~IEEE,}
        % <-this % stops a space
\thanks{Yukuan Zhang, Yunhua Jia, Housheng Xie, Mengzhen Li, Limin Zhao, Yang Yang and Shan Zhao are with the Laboratory of Pattern Recognition and Artificial Intelligence, Yunnan Normal University, Kunming, 650500, China (e-mail:\{{Ykuan$\textunderscore$Zhang, jia$\textunderscore$yunhua, xiehoushengsc, 18937346496, limin$\textunderscore$zhaom, yyang$\textunderscore$ynu}\}@163.com, szhaoynnu@yeah.net)}
\thanks{Corresponding authors:Yang Yang and Shan Zhao}}
%\markboth{IEEE TRANSACTIONS ON MULTIMEDIA}%
%{Shell \MakeLowercase{\textit{Yukuan~Zhang et al.}}: Rt-Track: Robust Tricks for Multi-Pedestrian Tracking}

\maketitle

\begin{abstract}
Object tracking is divided into single-object tracking (SOT) and multi-object tracking (MOT). MOT aims to maintain the identities of multiple objects across a series of continuous video sequences. In recent years, MOT has made rapid progress. However, modeling the motion and appearance models of objects in complex scenes still faces various challenging issues. In this paper, we design a novel direction consistency method for smooth trajectory prediction (STP-DC) to increase the modeling of motion information and overcome the lack of robustness in previous methods in complex scenes. Existing methods use pedestrian re-identification (Re-ID) to model appearance, however, they extract more background information which lacks discriminability in occlusion and crowded scenes. We propose a hyper-grain feature embedding network (HG-FEN) to enhance the modeling of appearance models, thus generating robust appearance descriptors. We also proposed other robustness techniques, including CF-ECM for storing robust appearance information and SK-AS for improving association accuracy. To achieve state-of-the-art performance in MOT, we propose a robust tracker named Rt-track, incorporating various tricks and techniques. It achieves 79.5 MOTA, 76.0 IDF1 and 62.1 HOTA on the test set of MOT17.Rt-track also achieves 77.9 MOTA, 78.4 IDF1 and 63.3 HOTA on MOT20, surpassing all published methods. %The code and model are available at: \url{https://github.com/../..}.

\end{abstract}


\section{Introduction}
\IEEEPARstart{T}{he} core of dialectical materialism is that ``All matter is continually moving and changing, and things are interconnected, permeate, and influence each other.'' Multi-object tracking tasks (MOT) have progressed rapidly in recent years, benefit from the substantial advancement of object detection tasks\citep{yolox,8060595,faster-rcnn,centernet}, in addition to playing a non-negligible role to dealing with motion blur problems in video object detection tasks \citep{9747993}.

MOT's goal is to preserve the identities of numerous objects in a continuous video sequence, which has significant application value in the fields of intelligent surveillance, intelligent transportation, and automatic driving. Maintaining continuous identity across subsequent video frames necessitates overcoming many challenges, such as (1) Tracking failure caused by occlusion, object densification. (2) The Kalman filter's error accumulation and trajectory drift are caused by the unstable detection box's input to the filter and the trajectory's unsuccessful matching. Therefore, the objectives of our research is to maintain accurate tracking in challenging situations like occlusion, object density, and small targets etc. how to avoids error brought on by unstable detection boxes, and extract distinguishing appearance features.
\begin{figure}[!t]	
	\centering
	\includegraphics[width=3.5in]{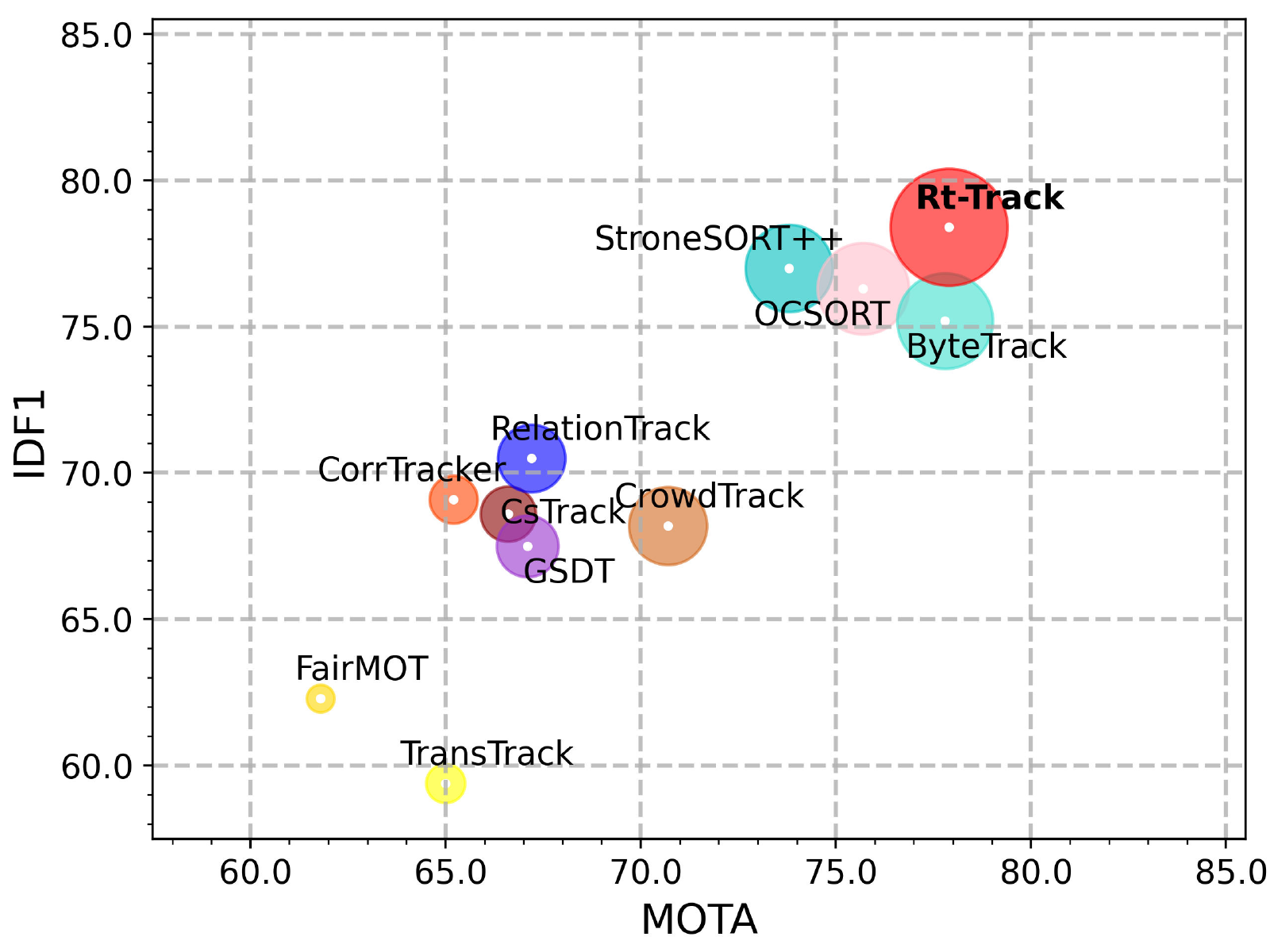}
	\caption{MOTA-IDF1 comparisons of different trackers on the test set of MOT20. Our Rt-Track achieves 77.9 MOTA, 78.4 IDF1 on MOT20 test set, outperforming all previous trackers. Details are given in Table \ref{tab_mot20}.}
	\label{duibi}
\end{figure}
Our motivation includes three points.

1) Existing state-of-the-art (SOTA) trackers, such as OC-SORT\citep{oc-sort}, have not considered the issue of trajectory smoothness when applying trajectory direction consistency during the association stage. Unsmooth trajectory will increase errors during the association stage.

2) SOTA approaches based on the DeepSORT\citep{deepsort} tracker extension, such as StrongSORT\citep{strongsort}, only take into account high-level semantic information when employing appearance feature modeling. Furthermore, in challenging scenarios, methods such as DeepSORT and StrongSORT may not have enough discriminative power when extracting object appearance information.

3) The mechanism of the appearance feature library is a crucial recovery strategy for the trajectory loss brought on by long-term occlusion. \citep{deepsort,trmt} remembers all consecutive feature sequences, which introduces more redundant information and increases memory consumption. Although \citep{botsort} saves computational overhead by remembering the fused feature sequences, it also incorporates more negative features.

We recognise the limitations of SOTA trackers. We find that by addressing these limitations, advanced performance can be achieved.

In this paper, we propose a robust skillful tracker, named Rt-Track. We take DeepSORT as the benchmark tracker, and propose corresponding innovations in motion and appearance models and matching strategies, It achieves robust tracking performance in complex scenarios. The proposed Rt-Track surpasses the SOTA tracker in most MOT metrics, that the results are shown in Fig.\ref{duibi}.

The main contributions of our work can be summarized as follows:
\begin{itemize}
	\item{\textbf{Smooth trajectory predicts direction consistency (STP-DC).} A simple and effective trajectory smoothing mechanism is designed and a new trajectory direction consistency calculation method is proposed to improve the matching accuracy.}
	\item{\textbf{Hyper grain feature extraction network (HG-FEN).} A more discriminative feature extraction network is designed to enhance the focus on the discriminative features of tracked objects.}
	\item{\textbf{Coarse EMA and Fine EMA embedded cluster mechanism (CF-ECM).} Design a new micro-appearance feature memory library mechanism, it can save the more robust appearance features of the tracked objects,  improved affinity metric accuracy, especially in occlusion situations. reduce memory consumption. Meanwhile save calculating time.}
	\item{\textbf{Skillful association strategy (SK-AS).} We combine three similarity decision methods (intersection-to-union ratio, orientation consistency, and appearance features) to construct a cost matrix and propose a new matching strategy to achieve higher tracking performance.}
\end{itemize}
\section{Related Work}
\subsection{Tracking Paradigm}
The current online multi-target tracking methods mainly follow the tracking-by-detection (TBD), joint-detection- association (JBT) and Transformer paradigms. The first step of TBD uses a high performance detector to locate the target location, and a tracking model is used for association in the second step. During the tracking phase, the mainstream association methods are classified into association-by-movement (ABM) and joint-movement-appearance (JMA) architectures. Comparatively, the ABM\citep{sort,oc-sort,bytetrack} method is faster than JMA\citep{strongsort,botsort,MOTDT} in terms of speed, but lower in tracking accuracy. 

TBD allows the detection and tracking algorithms to be trained separately such that optimal models are obtained for each of the two tasks, and thus the paradigm is able to achieve higher tracking accuracy. JBT improves tracking efficiency by unifying detection and tracking algorithms trained in a single network\citep{centertrack,fairmot,corr-track} to obtain tracking information in video sequences. However, maintaining a balance between the detection and tracking components during training is a challenging task that affects the model's ability to achieve higher tracking accuracy. 
Recently, more and more researchers have been introducing transformers and multi-head attention mechanisms into visual multi-object tracking tasks\citep{RelationTrack,TransCenter}. This is due to the success of these techniques in natural language processing tasks and their effectiveness in learning long-term dependencies in sequences.

We believe that TBD is still the most popular tracking paradigm, which is able to balance time and efficiency.

\subsection{Motion model}
In MOT, the Kalman filter\citep{kalman1960contributions} (KF) that assumes linear motion\citep{sort,deepsort,trmt,strongsort,bytetrack,fairmot} is often used to model the motion information of objects. SORT\citep{sort} achieves well performance in simple scenarios by only using the detection box and predicted box from the KF to calculate the IOU (Intersection-over-Union) distance. ByteTrack\citep{bytetrack} has designed high and low scoring matching steps and uses IOU associations at each stage. It enhances the tracking performance of non-linear moving objects. The TGCN\citep{tgcn} uses the optimal state estimate of the KF to calculate the directional consistency of the trajectory, in order to handle the target identity switching problem in relative driving. OC-SORT\citep{oc-sort} suggests a direction consistency calculation approach centered on the observation value in order to prevent the state noise accumulation of the KF during occlusion. TraDes\citep{trades}, CenterTrack\citep{centertrack} uses Centernet as the benchmark framework\citep{centernet}, adding a head branch to predict the trajectory offset map, which is used to handle associations after long periods of occlusion.

We recognize that the motion model has not been fully developed, for example, IOU is difficult to perform well in complex scenarios, and the directional-consistency-association-strategy (Dc-AS) is easily affected by the smoothness of the trajectory. Unlike the aforementioned approach, we have designed a smoothness strategy to optimize the input and output state variables of KF, making Dc-AS more accurate in associating in complex scenarios.
\subsection{Appearance model}
Modelling the appearance of objects enables longer tracking. It is able to recover the identity of a target that has been lost for a long time. Patch-based feature extraction methods\citep{fastreid,zheng2017discriminatively,hermans2017defense,bot} are a standard scheme adopted by multi-target trackers\citep{poi-track,mahmoudi,zhu2018online,deepsort,botsort,strongsort}. Its key challenge is to extract discriminative features from the patch. 

DeepSORT\citep{deepsort} uses a conventional convolutional neural network (CNN) to extract 128-dimensional features. The feature extraction network is quick, but its discriminative power is poor in practical situations. With the rapid development of the field of pedestrian re-identification (Re-ID) in recent years, existing SOTA trackers, such as \citep{strongsort,botsort}, use the Re-ID model in the Fastreid library\citep{fastreid} when extracting appearance features. In comparison to typical CNN, this technique extracts a feature embedding vector with a dimension of 2048, which has higher robustness for tracking objects in complicated scenarios.

Although the feature extraction methods in the \citep{fastreid} have achieved strong performance, we think that when applying Re-ID tasks to MOT, more complex situations need to be considered, such as dense crowds and occlusions. We use the BOT\citep{bot} model from the Fastreid library, with a ResNet-50\citep{resnet} backbone network, the core network is upgraded somewhat to meet the difficulty of tracking multiple objects simultaneously in complex scenarios.
\begin{figure*}[t]	
	\centering
	\includegraphics[width=7.0in]{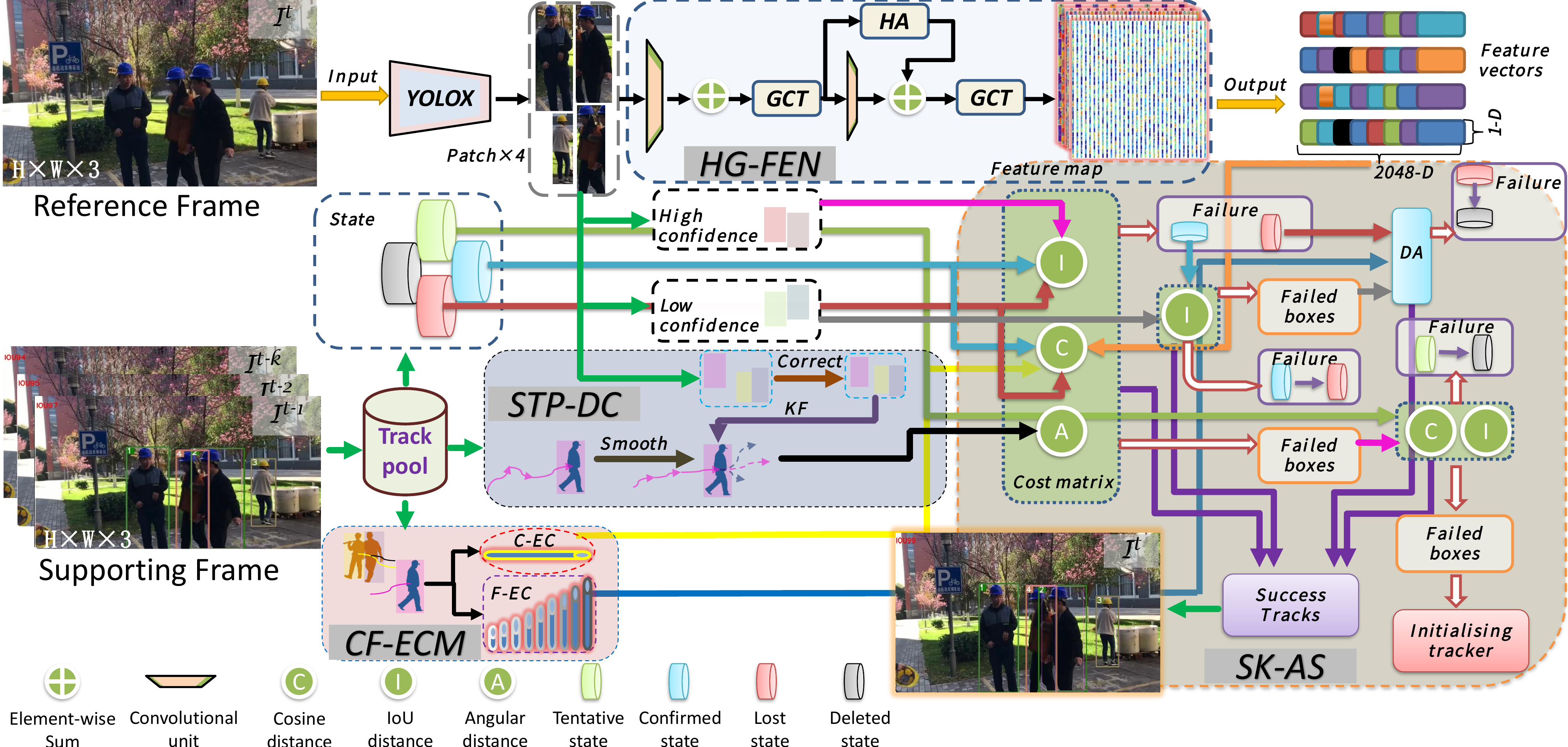}
	\caption{Diagram of the network structure of Rt-Track. The diagram also shows the internal structure of our proposed four robustness tricks.}
	\label{Rt-Track}
\end{figure*}
\section{Proposed Method}
In this section, we provide a detailed description of the four modifications and improvements that we propose, which are based on the classical tracker DeepSORT. The pipeline of our algorithm is present in Fig.\ref{Rt-Track}.
\subsection{STP-DC}
In complex scenarios, the measurement vectors of the KF can be disturbed by the surrounding environment, leading to uncertain positioning. After prolonged occlusion, the noise of the KF system grows exponentially\citep{oc-sort}, resulting in drifting trajectories.

In order to efficiently model motion in complex scenes, we propose a smooth trajectory prediction direction-consistent computational strategy (STP-DC) to overcome the positioning disturbance of KF measurement vectors and the drifting of optimal state vectors. We find that by modifying the KF system variables, we can obtain smoother trajectories and use this smooth trajectory for Dc-AS to achieve higher tracking performance. Details and KF system settings can be found below.
\begin{equation}\label{KF's sate}
	\bm{x(n)} = [x_{c}(n),y_{c}(n),\dot{x}_{c}(n),\dot{y}_{c}(n)]^{T}
\end{equation}
\begin{equation}\label{KF's meas}
	\bm{z(n)} = \begin{cases}
		[B_x,B_y]^T,if\ n<k \\
		[z_{x_{c}}(n),z_{y_{c}}(n)]^{T},otherwise
	\end{cases}
\end{equation}

We choose to define the KF state vector as in Eq.(\ref{KF's sate}) to represent the state of each object centroid trajectory in a sequential frame. Eq.(\ref{KF's sate}) contains four tuples, where ($x_{c}$,$y_{c}$) represents the 2D coordinates of the object centroid in the image patch.($\dot{x}_{c}$,$\dot{y}_{c}$) denotes the velocity vector corresponding to the center point coordinates. In order to obtain smoother trajectories and thus avoid tracking drift due to the accumulation of KF errors, we correct the traditional KF measurement vector. 
\begin{equation}
	\label{KF's correct meas}
	\begin{split} %cases
		\displaystyle z&_{x_{c}}(n) = O_{x}+\frac{\parallel{\bm{OA}}\parallel_{2}+\parallel{\bm{OB}}\parallel_{2}}{2}cos\{\frac{\pi}{180}\times[arccos(\\
		&\
		\frac{\bm{OO'\cdot\bm{OA}}}{\parallel{\bm{OO'}}\parallel_{2}\parallel{\bm{OA}}\parallel_{2}})+\frac{1}{2}arccos(\frac{\bm{OA\cdot\bm{OB}}}{\parallel{\bm{OA}}\parallel_{2}\parallel{\bm{OB}}\parallel_{2}})]\}\\[5pt]
		\displaystyle z&_{y_{c}}(n) = O_{y}+\frac{\parallel{\bm{OA}}\parallel_{2}+\parallel{\bm{OB}}\parallel_{2}}{2}sin\{\frac{\pi}{180}\times[arccos(\\
		&\
		\frac{\bm{OO'\cdot\bm{OA}}}{\parallel{\bm{OO'}}\parallel_{2}\parallel{\bm{OA}}\parallel_{2}})+\frac{1}{2}arccos(\frac{\bm{OA\cdot\bm{OB}}}{\parallel{\bm{OA}}\parallel_{2}\parallel{\bm{OB}}\parallel_{2}})]\}
	\end{split}
\end{equation}

Our KF measurement vector as in Eq.(\ref{KF's meas}), where ($z_{x_{c}}(n),z_{y_{c}}(n)$) denotes the corrected 2D coordinates. $B\in[B^0,B^1,...,B^n]$ is the bounding box center output by the detector. $z_{x_{c}}(n)$, $z_{y_{c}}(n)$ can be obtained by Eq.(\ref{KF's correct meas}), of which $O$, $A\in M=\{[F^0,F^1,...,F^k],[U^{k+1},U^{k+2},...,U^n]\}$, with $U$,$F$ represent the optimal object centroid position of the KF output as in Eq.(\ref{KF's update process}) and the centroid position after linear fitting as in Eq.(\ref{KF's fit point}), respectively. $O'$ is the point of $O$ on the ray in the specified direction.
\begin{equation}\label{KF's Q and R}
	\begin{split}
		\displaystyle \bm{Q} &= diag(\sigma_{qx}^2,\sigma_{qy}^2,\sigma_{vx}^2,\sigma_{vy}^2)\\
		\displaystyle \bm{R} &= diag(\sigma_{rx}^2,\sigma_{ry}^2)
	\end{split}
\end{equation}
% \displaystyle 用于使得两行公式不拥挤

We choose to define the metric $\bm{Q},\bm{R}$ as shown in Eq.(\ref{KF's Q and R}), where $\sigma$ is the noise factor corresponding to each component in $\bm{x(n)}$ and $\bm{z(n)}$, with $\sigma_{qx}^2=\sigma_{qy}^2=\sigma_{vx}^2=\sigma_{rx}^2=1$, $\sigma_{vy}^2=0.01$, $\sigma_{ry}^2=10$, i.e. , process noise covariance $\bm{Q}$ and observation noise covariance $\bm{R}$, they are time indepent.

We follow the Recursion idea of the traditional KF, in a time step it has a prediction process as in Eq.(\ref{KF's prediction process}) and an update process as in Eq.(\ref{KF's update process}), in which $\bm{A}\in\mathbb{R}^{4\times4}$ is the state transition model, $\bm{P}\in\mathbb{R}^{4\times4}$ is the estimated covariance matrix of the state vector $\bm{x(n)}$ in the prediction and update process of KF, $\bm{K}\in\mathbb{R}^{4\times2}$ is Kalman Gain, $\bm{H}\in\mathbb{R}^{2\times4}$ is the observation model. In our method $z_n=\bm{z(n)}$ and $U^n=(\hat{x}_{nx},\hat{x}_{ny})$.
\begin{equation}\label{KF's prediction process}
	\begin{split}
		\displaystyle \hat{x}^-_n &=\bm{A}\hat{x}_{n-1} \\
		\displaystyle \bm{\hat{P}}^-_n &=\bm{A\hat{P}}_{n-1}\bm{A}^T + \bm{Q}
	\end{split}
\end{equation}
\begin{equation}\label{KF's update process}
	\begin{split}
		\displaystyle \bm{K}_n &= \frac{\bm{\hat{P}}^-_n\bm{H}^T}{\bm{H\hat{P}}^-_n\bm{H}^T + \bm{R}}\\
		\displaystyle \hat{x}_n &= \hat{x}^-_n + \bm{K}_n(z_n - \bm{H}\hat{x}^-_n)\\
		\displaystyle \bm{\hat{P}}_n &=(\bm{I}-\bm{K}_n\bm{H})\bm{\hat{P}}^-_n
	\end{split}
\end{equation}

During the initial time, the real target in the sequence has a regular state. However, the disturbance of the target detection boxes affects the smoothness of the KF output trajectories. As a result, after k occurrences of the object, we take $[U^0,U^1,...,U^k]$ as the base point and fit the function $F(x) = ax + b$ using non-linear least squares. Then, we use Eq.(\ref{KF's fit point}) to map $U^*$ to $f(x)$ to obtain $F^*$.
\begin{equation}\label{KF's fit point}
	\begin{split}
		\displaystyle F^*_x &= (aU^*_y + U^*_x -ab)(a^2 + 1)^{-1} \\
		\displaystyle F^*_y &= (a^2U^*_y + aU^*_x + b)(a^2 + 1)^{-1}
	\end{split}
\end{equation}
\begin{figure*}[!ht]	
	\centering	
	\subfloat[DeepSORT]
		{\includegraphics[width=2in]{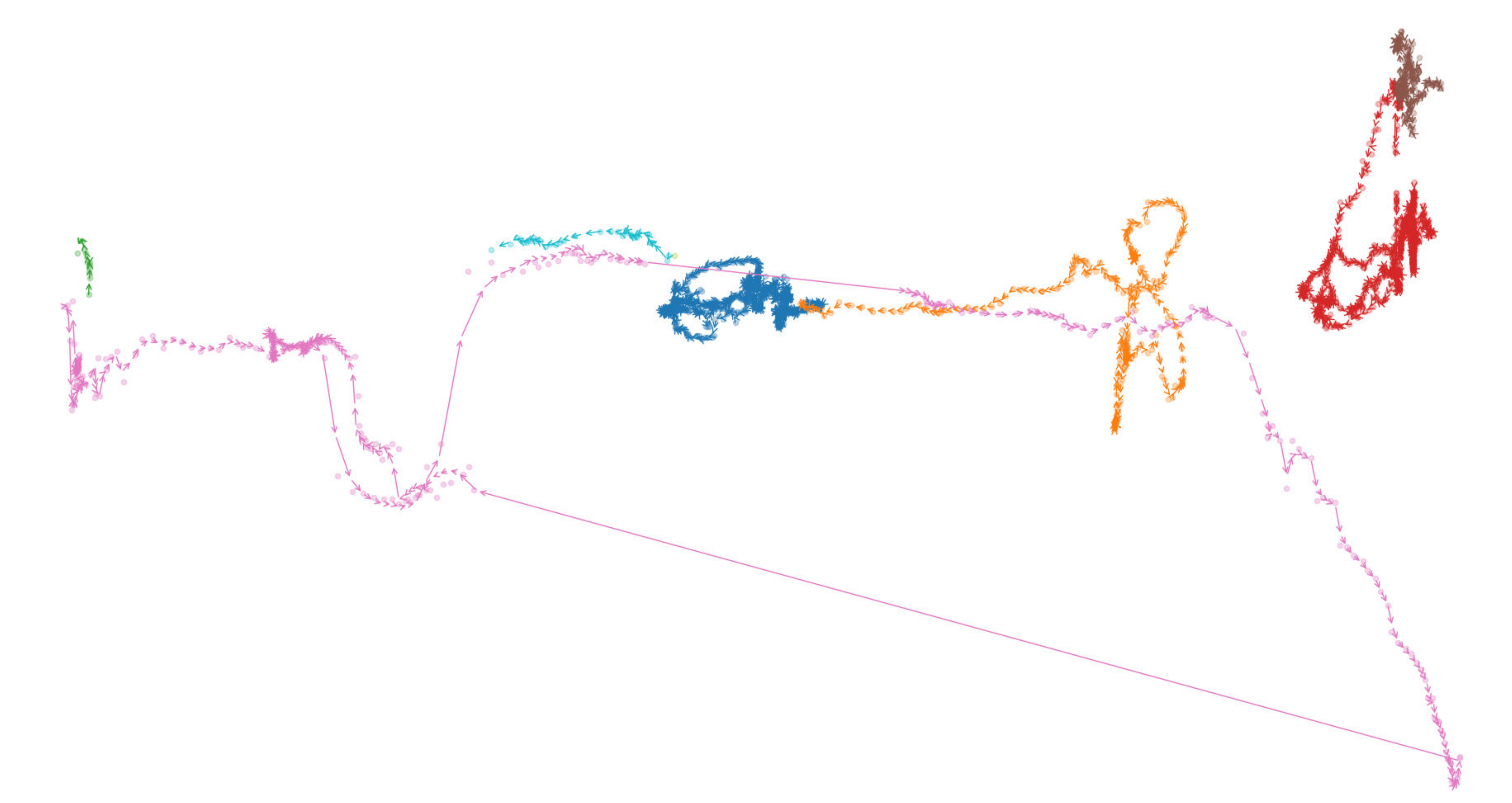}}
	\label{fig11}
	\hfil
	\subfloat[The proposed trajectory smoothing method]
		{\includegraphics[width=2in]{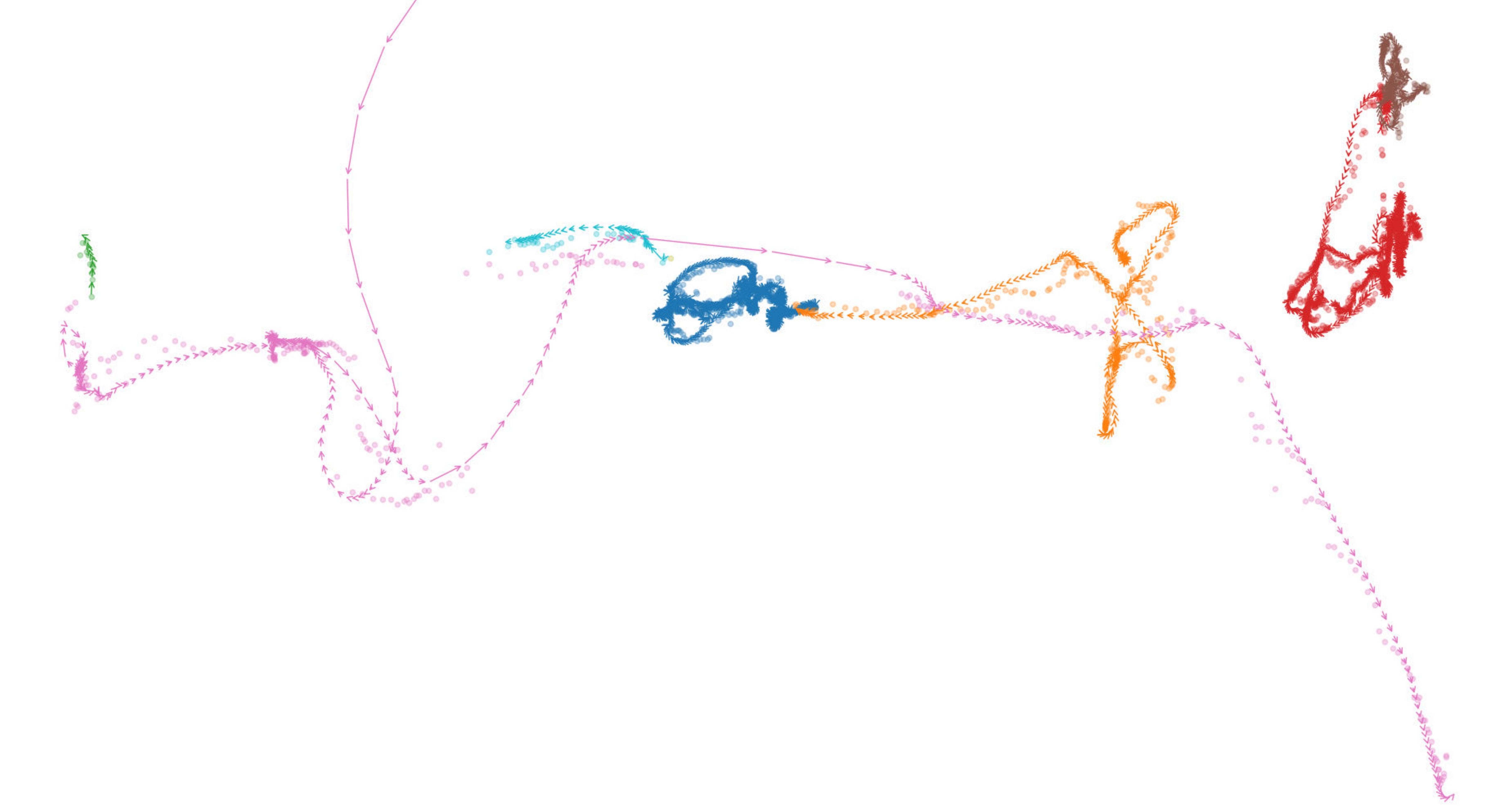}}
	\label{fig12}
	\hfil
	\subfloat[Our STP-DC fusion with DeepSORT's matching strategy]
		{\includegraphics[width=2in]{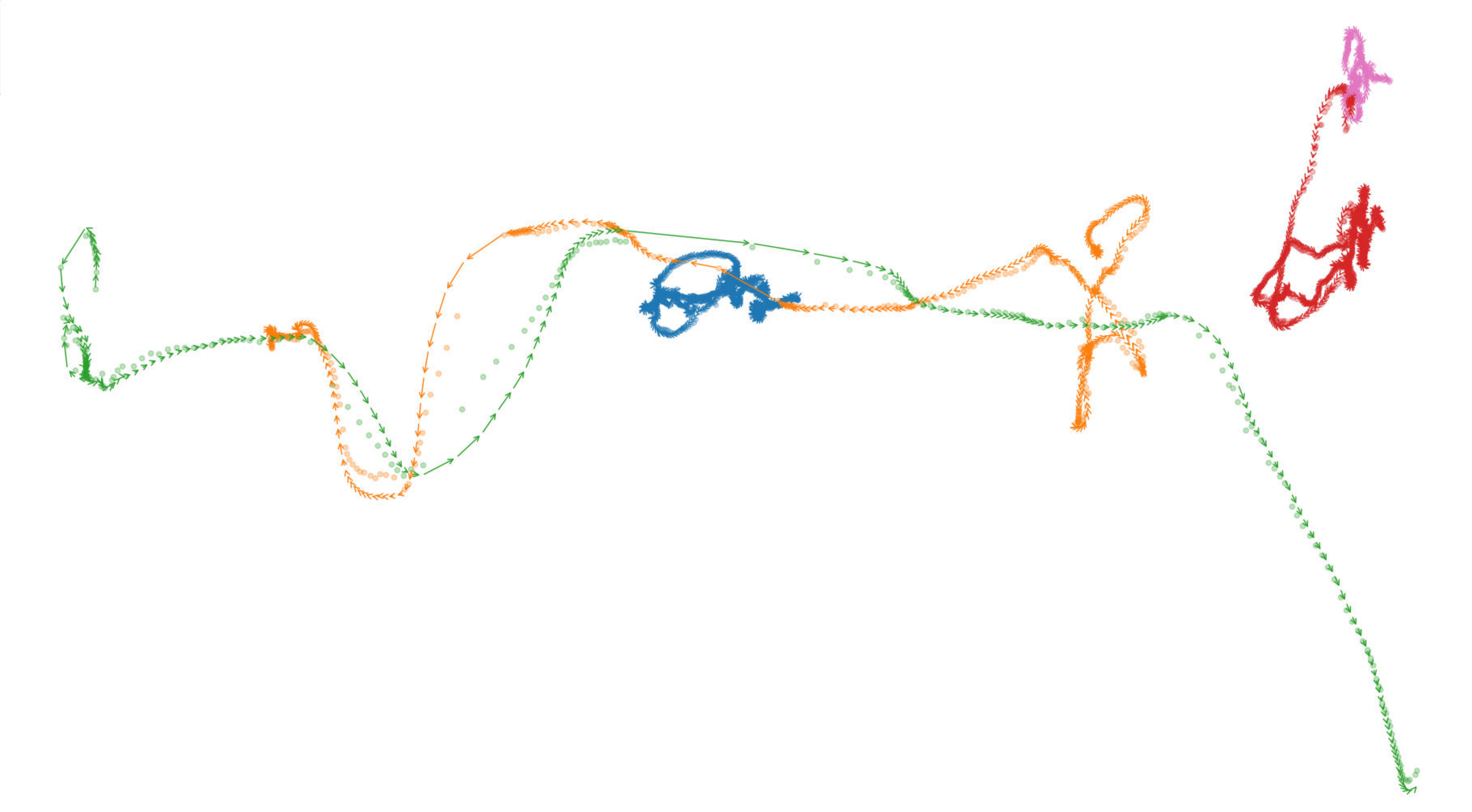}}
	\label{fig13}
	\caption{Multi-target motion trajectory map. The arrows($\uparrow$) in the graph indicate the direction of the object's motion. The transparent points($\bullet$) in the (a,b) graph represent the center $(B_x,B_y)$ of the bounding box output by the detector, and in the (c) graph represent the center $(z_{x_c},z_{y_c})$ of the corrected detection bounding box. The correct tracking of all objects is achieved in the bottom graph, where the \textcolor[rgb]{0.121568,0.46667,0.705882}{blue} represents the track identity of the occluded object, the \textcolor[rgb]{0.211765,0.6392157,0.211765}{green} and \textcolor[rgb]{1,0.498,0.0549}{yellow} represent the track identifiers via of the occluded object.}
	\label{fig1}
\end{figure*}
\indent
In Fig. \ref{fig1}(b), we can find that our method obtains a smoother trajectory map compared to the traditional DeepSORT as Fig. \ref{fig1}(a). To further implement the Dc-AS, we propose a new associations strategy called STP-DC, as shown in Eq.(\ref{Directional consistency calculation}).
\begin{equation}\label{Directional consistency calculation}
	\begin{split}
		\displaystyle C_d &= \{Z,\mathbb{T},X\}_{m\times{n}}\\
		&=\frac{\pi}{180}arccos\{\frac{[\bm{ \tau_j}^{(M_{n-1})}\bm{x_i}]\cdot[\bm{\tau_j}^{(M_{n-3})}\bm{\tau_j}^{(M_{n-1})}]}{\parallel{\bm{ \tau_j}^{(M_{n-1})}\bm{x_i}}\parallel_{2}\parallel{\bm{\tau_j}^{(M_{n-3})}\bm{\tau_j}^{(M_{n-1})}}\parallel_{2}}\}\\	
	\end{split}
\end{equation}

Where $Z = \{z_1,z_2,...,z_n\}$, $\mathbb{T}=\{\tau_1,\tau_2,...,\tau_m\}$ is the set of all activation state trackers at the current time, and $X= \{x_1,x_2,...,x_n\}$, with $x_i \in \mathbb{R}^m$ is the result of all trackers updated by the KF at the current time. Fig. \ref{Directional consistency calculation}(c) validates the effectiveness of STP-DC, which is described in the experimental section of this paper.
\subsection{HG-FEN}
It is also know that the bottom layer of the CNN is able to capture more detailed information, i.e., color, texture, edges, and angles. As the number of downsampling increases and the perceptual field gradually expands, the deeper layer is able to extract rich semantic features. However, due to the reduction in resolution of the high-level convolutional feature map, small samples of pedestrians cannot be effectively represented\citep{zhang2016faster}. 

In crowded scenes, significant occlusion exists between targets. However, patch-based feature extraction methods extract background information in addition to the target. We believe that focusing on the shallow features of the targets in complex scenes can enhance intra-class compactness and expand inter-class differences.
\begin{figure}[h]	
	\centering
	\includegraphics[width=3.5in]{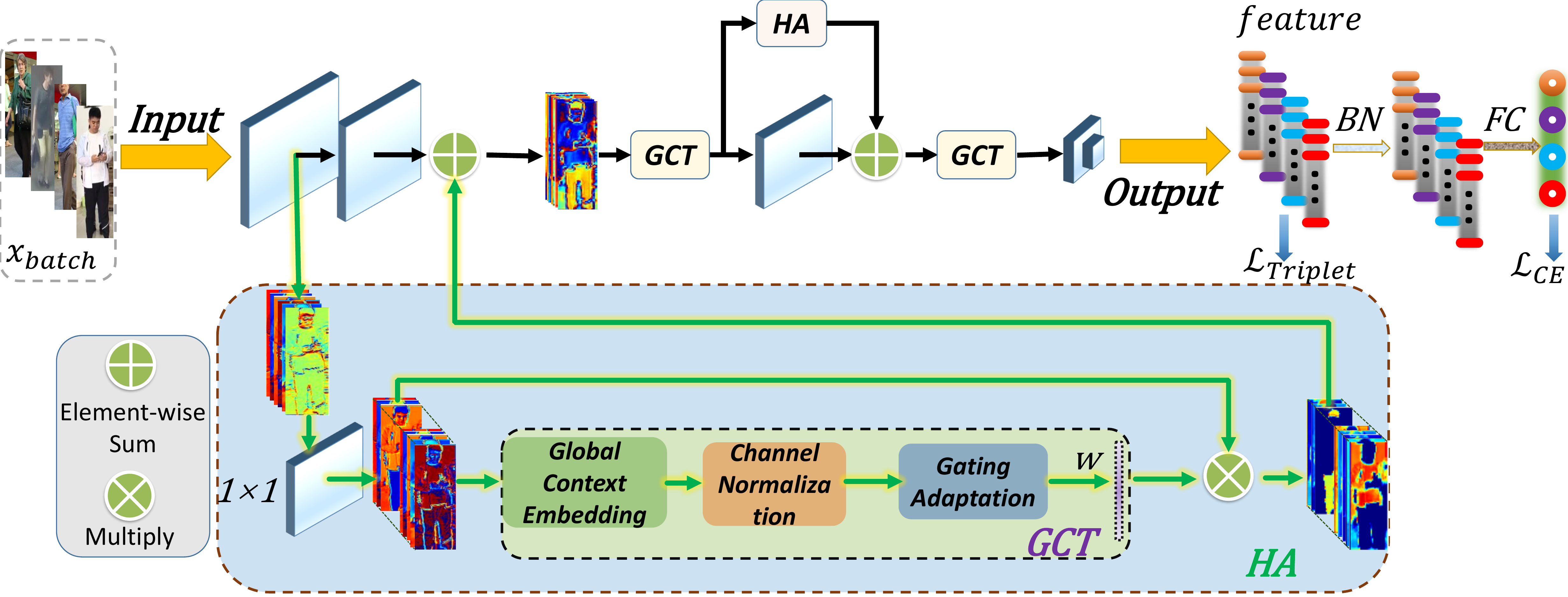}
	\caption{our HG-FEN network architecture. HA stands for Hyper-grain attention Module. }
	\label{featuer_network}
\end{figure}

In our implementation, we use ResNet-50\citep{resnet} as the backbone network and design the hyper-grain attention (HA) module to extract more robust appearance features, our network architecture is shown in Fig.\ref{featuer_network}.

We adopt a $1\times1$ kernel in the HA module to resize the shallow feature map, which is used to fuse the fine-grained information of the shallow features with the coarse-grained information of the deep features.

To obtain more discriminative target features in patch, we use a transform unit GCT\citep{gct} to model the channel relevance and contextual information of the fused feature maps, which is used to make the network output a hyper-grain feature map for each batch of patches.

In our method, HA module is added on top of in stages conv2, conv3, and conv4. The network input size is $x_{batch} \in\mathbb{R}^{b\times c\times h\times w}$. The output embedding map dimension after the fully connected layer is $e(x) \in\mathbb{R}^{2048}$. In the tracking step, we crop the $x_{batch}$ of each object based on the detection response, and then use the cosine distance to measure the affinity between the embedding pairs. To adapt to the MOT task, we use the first half of MOT17\citep{mot17} and MOT20\citep{mot20} as the training set. Similar to \citep{bot}, we jointly utilize the triplet loss with label smoothing\citep{ls} (LS) and cross-entropy loss in different feature spaces, as shown in Eq.(\ref{Triplet loss}) and Eq.(\ref{Cross-entropy loss}).
\begin{equation}\label{Cross-entropy loss}
	\begin{split}
		q_i &= \begin{cases}
			1 - \frac{N - 1}{N}\varepsilon, if\ i=y \\
			\varepsilon / N, otherwise
		\end{cases} \\
		\mathcal{L}_{CE} &= \sum_{i=0}^{n}-q_ilog(p_i)\begin{cases}
			q_i=0,y\ne i \\
			q_i=1,y=i
		\end{cases}	
	\end{split}
\end{equation}
\begin{equation}\label{Triplet loss}
	\mathcal{L}_{Triplet}=\sum_{i=0}^{n}[\left\|e(x_i^a)-e(x_i^p) \right \|^2_2-\left\|e(x_i^a)-e(x_i^n) \right \|^2_2+\alpha]_+
\end{equation}

where denote $y$ as truth label and $p_i$ as prediction logits of class $i$. $\varepsilon$ is set to be $0.1$. The $\mathcal{L}_{CE}$ maps the distribution of features of each class across different subspaces, while the $\mathcal{L}_{Triplet}$ ensures that the cosine distances between embedding of the same identity are small.

We evaluate our model on Market1501\citep{Market1501}, See Table \ref{tab4}, almost without additional computational complexity, with achieve a Rank-1 , mAP and mINP of 94.8, 87.1 and 62.5, respectively, an improvement of 0.5, 0.8 and 1.3 compared to the benchmark\citep{bot}.
\subsection{CF-ECM}
\begin{figure}[!h]	
	\centering
	\includegraphics[width=3.2in]{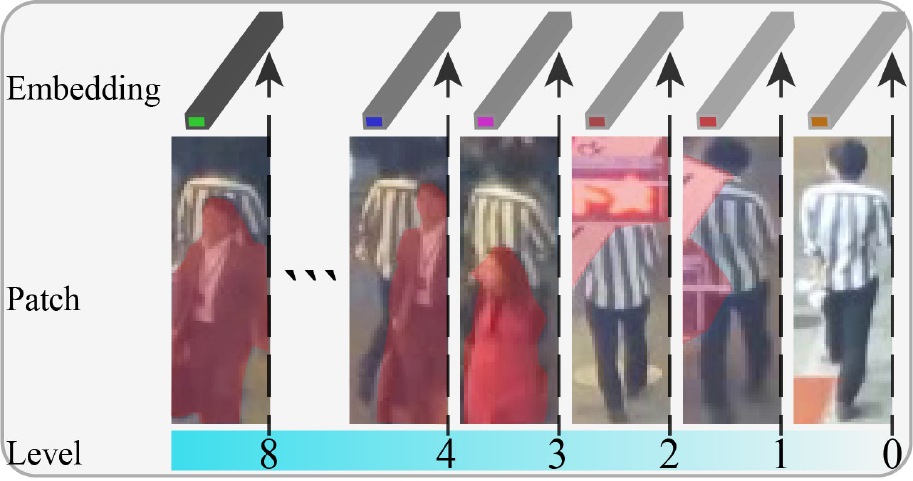}
	\caption{F-EMA.Occlusion levels are classified according to the confidence level of object detection. Each box in the figure possesses a different level of occlusion. The \textcolor[rgb]{0.77,0.26,0.31}{mask} represents the occluded area.}
	\label{F-FMA}
\end{figure}

The embedding cluster is used to store previous appearance information and it is able to reactivate lost tracks to achieve higher IDF1. In consecutive sequence frames, the motion range of the targets remains within a small range. Therefore, earlier methods\cite{deepsort} are prone to preserving redundant information, and early trajectory information is lost over time. 

In recent years, researchers have applied a approach to overcome the limitations of earlier methods. They\citep{trmt,botsort,strongsort} use exponential moving average\citep{ema} (EMA) to update the previous information of trajectories. However, the changes in targets in complex scenes are diverse. As the degree of occlusion increases, the advantage of affinity measures based on saved information may weaken in complex scenes.

We propose a novel micro embedding group mechanism that can preserve richer robustness information along the temporal dimension. It constructs Coarse-EMA (C-EC) and Fine-EMA embedding clusters (F-EC) based on the level of occlusion. 

We update the C-EC using Eq.(\ref{coarse EMA}), similar to the approach used in \citep{strongsort}. This reduces the time consumption. 
\begin{equation}\label{coarse EMA}
	e^n_i=\alpha e^{n-1}_i+(1-\alpha)f^n_i
\end{equation}
\begin{equation}\label{fine EMA}
	\begin{split}
		\{e^n_i\}_{Conf}&=\alpha \{e^{n-1}_i\}_{Conf}+(1-\alpha)\{f^n_i\}_{Conf},\\
		Conf&\in \{(0.1,0.2],(0.2,0.3],...,(0.9,1)\}
	\end{split}
\end{equation}

In the F-EC, we classify occlusion into 9 levels (shown in Fig. \ref{F-FMA}) and use Eq. (\ref{fine EMA}) to update it. 
This updates the appearance state ${e^n_i}_{Conf}$ of embedded clusters at the level where $Conf$ is located for the $i$-th tracklet at frame $n$. In Eq. (\ref{fine EMA}), ${f^n_i}_{Conf}$ refers to the appearance information of the detection bounding box with confidence $Conf$ that matches the tracklet. A momentum term with $alpha = 0.9$ is used. 

F-EC provides a finer expression of appearance for objects with different levels of occlusion, and can be used for appearance affinity calculation in complex scenarios. 

CF-ECM can fully leverage the advantages of appearance features in MOT tasks while saving computing resources. When combined with motion models, it can achieve higher tracking accuracy. Our CF-ECM has been effectively validated in the ablation study section, and the pseudo-code for F-EC is shown in Algorithm \ref{alg:alg1}.
\begin{algorithm}[t]
	\caption{Pseudo-code of F-EC.}\label{alg:alg1}
	\begin{algorithmic}[1] 
		\STATE \textbf{Input:}Matching trajectory-detecting pairs $\mathcal{M(\mathcal{T},\mathcal{D})}$;The\\ \hspace{0.5cm} update level $\mathcal{I}$ of the F-EC embedding cluster\\ \hspace{0.5cm} corresponding to $\mathcal{T}.$
		\STATE \texttt{/* Initialize the feature pool for each trajectory */}
		\STATE Initialization: $\mathcal{P} \gets \mathbf{0}$
		\STATE \textbf{if} \  $\mathcal{I}$ \  $is$  \ $none$ \ \textbf{then}
		\STATE \hspace{0.5cm}$i \gets 0.1$ 
		\STATE \hspace{0.5cm}\textbf{for} \ $t$ \ $in$ \ $\mathcal{T}$ \ \textbf{do}
		\STATE \hspace{1.0cm}\textbf{if} \  $ 0.1 + i < t.Conf < 0.2 + i$ \  \textbf{then}
		\STATE \hspace{1.5cm}$\mathcal{I} \gets 10 \times (0.1+i)-1$
		\STATE \hspace{1.5cm}$\mathcal{P}[\mathcal{I}] \gets t.feature \ / \ \left\|t.feature\right\|$
		\STATE \hspace{1.0cm}\textbf{end if}
		\STATE \hspace{1.0cm}$i \gets i+0.1$
		\STATE \hspace{0.5cm}\textbf{end for}
		\STATE \texttt{/* update F-EMA */}
		\STATE \textbf{else}
		\STATE \hspace{0.5cm}$\mathcal{D}.feature \gets \mathcal{D}.feature \ / \ \left\|\mathcal{D}.feature\right\|$
		\STATE \hspace{0.5cm}$\mathcal{T}.\mathcal{P}[\mathcal{I}] \gets \alpha\mathcal{T}.\mathcal{P}[\mathcal{I}]+(1-\alpha)\mathcal{D}.feature$ 
		\STATE \hspace{5.5cm}\texttt{// Eq.(\ref{fine EMA})}
		\STATE \hspace{0.5cm}$\mathcal{T}.\mathcal{P}[\mathcal{I}] \gets \mathcal{T}.\mathcal{P}[\mathcal{I}] \ / \ \left\|\mathcal{T}.\mathcal{P}[\mathcal{I}]\right\|$
		\STATE \textbf{end if}
		\STATE \textbf{Return:} $\mathcal{P}$
	\end{algorithmic}
	\label{alg1}
\end{algorithm}
\subsection{SK-AS}
The core idea of cascaded matching is to prioritize the matching of the tracklets that have been successfully associated most recently. We have designed a new deep-based association (DA) strategy, which differs from the previous approach\citep{deepsort}. We refer to the number of successful matches as positive depth $Pd_+$ and the opposite as negative depth $Nd_-$. We believe that trajectories with higher depth values in the reference frame should be prioritized in matching, as they carry more valuable supporting frame information compared to newly appearing trajectories. Giving them more attention in the MOT task can lead to better performance. The specific method is described as follows. 

\begin{figure}[!h]	
	\centering
	\includegraphics[width=3.5in]{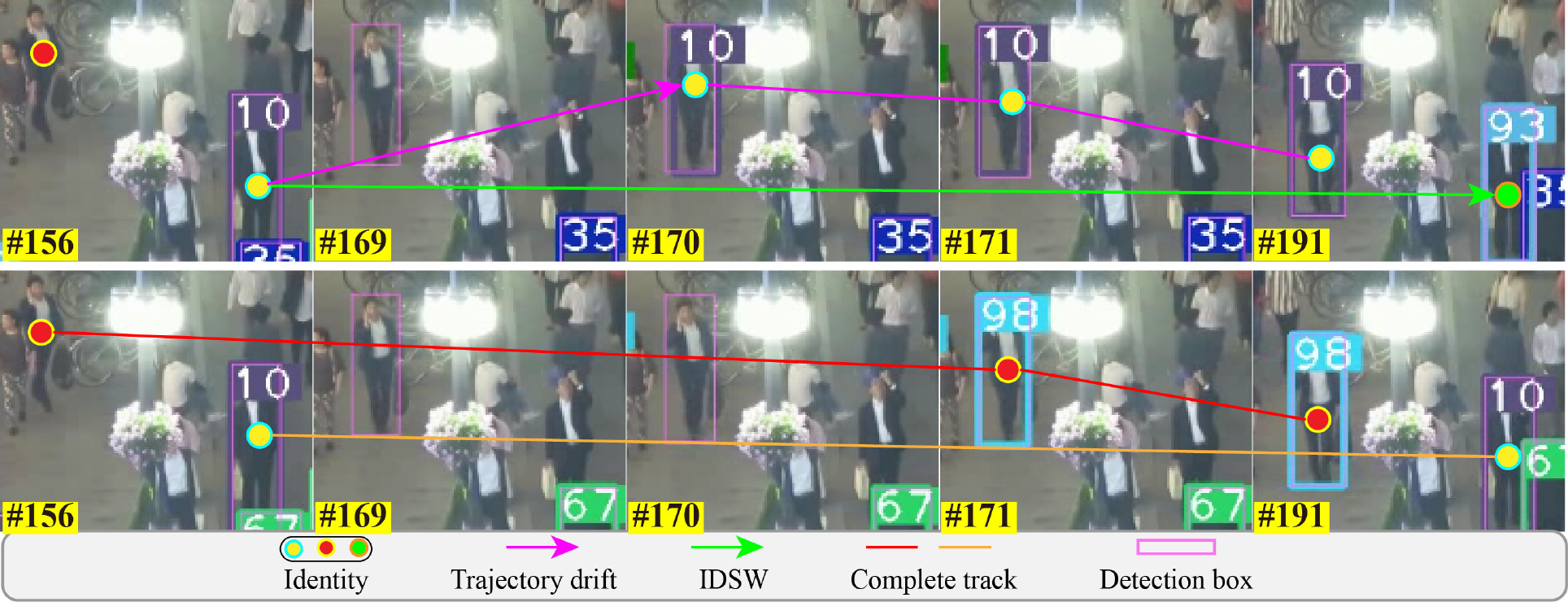}
	\caption{Comparison of matching methods. The top is the result of cascade matching and the bottom end is the result of our proposed DA algorithm. It can be noticed that the DA algorithm is able to avoid tracking errors.}
	\label{da}
\end{figure}

The affinity calculation is performed first for trajectories with $Pd_+ > 3Nd_-$ in the matching phase, followed by $Pd_+>2Nd_-$ and finally $Pd_+>Nd_-$. At each stage of the associated subproblem the trajectory-detection pair is solved using the Hungarian algorithm\cite{kuhn1955hungarian}. In Fig. \ref{da}, it can be observed that the tracking accuracy is improved with DA compared to the methods in \citep{deepsort}.

For accurate tracking, we fuse the motion and appearance models to construct the cost matrix $C$. We use the IoU distance metric $C_{iou}$ and the directional consistency metric $C_d$ (as defined in Eq. (\ref{Directional consistency calculation})) to calculate the positional affinity between trajectory-detection pairs, and the cosine distance metric $C_{cos}$ to measure the appearance affinity.

In contrast to the Eq. (\ref{weight sum cost}) fusion method, it uses a weighted sum form to construct the $C$. We adopt Eq. (\ref{our weight sum cost}) to construct ours cost matrix $C$. Where $C^{(i,j)}$ is the $(i,j)$ element of $C$. $C^{(i,j)} _{iou}$ is the IoU distance between tracklet $i$-th predicated bounding box and the $j$-th detection bounding box. $C^{(i,j)} _{cos}$ is the cosine distance cost between the CF-ECM of the $i$-th tracklet in the reference frame and the $j$-th detection patch in the support frame. $C^{(m,n)} _{d}$ is the angular distance between the $m$-th tracklet and the $n$-th tracklet.
\begin{equation}\label{weight sum cost}
	C = \lambda_1C_{iou} + \lambda_2C_{d} + (1-\lambda_1-\lambda_2)C_{cos}
\end{equation}

Where weight factor $\lambda_{1,2}$ is usually set to between 0 and 1.

Before generating $C^{(m,n)} _{d}$ and $C^{(i,j)} _{cos}$, we use IoUs as their threshold distance to reject unlikely pairs of tracklets and detections. Here the gate-threshold value is set to 0.5.
\begin{equation}\label{our weight sum cost}
	C^{(i,j)} = \lambda min\{C^{(i,j)} _{iou},C^{(i,j)} _{cos}\} + (1-\lambda)C^{(m,n)} _{d}
\end{equation}

To construct $C$, we first quantize the IoU distance metric and the cosine distance metric, then calculate the minimum of the two, similar to \citep{botsort}, and finally weight the sum of their results with the STP-DC. During the association phase of our tracker, Rt-track based on Eq. (\ref{our weight sum cost}) and the Hungarian algorithm to solve the linear assignment problem. To obtain a robust tracker, we combine our innovations with the final association task to form the skillful association strategy (SK-AS).

\section{Experiments}
\subsection{Datasets and Evaluation Metrics}
\textbf{Datasets.} To evaluate the robustness of the proposed tracker Rt-Track in MOT, we conducted experiments in the most popular MOTChallenge\citep{mot17,mot20} with the MOT17 and MOT20 datasets under the "private detection" protocol. MOT17 video sequences for challenges in static and moving cameras scenes, while MOT20 is used in crowded scenes. For ablation studies, we follow\citep{botsort,bytetrack}, by using the first half if each video in the training set of MOT17 for training and the last half for validation. Both datasets of the training set are also utilized to train the Re-ID model HG-FEN. 

\textbf{Metrics.} We use multiple object tracking accuracy(MOTA), ID F1 score(IDF1), higher order tracking accuracy(HOTA) as our main metrics, and other metrics include FP, FN, MT, ML, Rell, AssA, IDSW, FPS, etc., to evalute different aspects of the detection and tracking performance. MOTA\citep{mota} is computed based on FP, FN and IDSW, and focuses more on detection performance. IDF1\citep{idf1} evaluates the identity preservation ability and focus more on the association performance. HOTA\citep{hota} explicitly balances the effect of performing accurate detection, association, and localization into a single unified metric.
\subsection{Implementation details}
All the experiments are implemented using PyTorch and run on a desktop with 11th Gen Intel(R) Core(TM) i7-11700K @ 3.60GHz and a single NVIDIA GeForce RTX 3090 GPU. we directly apply the publicly available detector of YOLOX\citep{yolox} trained by\citep{bytetrack} for MOT17, MOT20, and ablation study on MOT17. For the feature extractor, HG-FEN with ResNet-50 as the backbone and Imagenet-pretrained model\citep{imagenet} as the initialized weights. We train our HG-FEN model on MOT17 and MOT20, its parameters are updated using the Adam optimizer\citep{adam} with weight decay of $5\times10^{-4}$. During the training procedure, the initial learning rate is $3.5\times10^{-4}$, input batch size is set as 64 and the resolution of every image is $256\times128$. Total training 120 epochs.

For the tracker, we set four tracklet states, including tentative, confirmed, deleted and lost state. To initialize a new tracklet, in the first frame of each video sequence we set the tracklet with a confidence score larger than the new tracklet threshold of 0.7 to the confirm state, while subsequent frames in the case where the condition is met will be set to the tentative. For the lost tracklets, we keep them for 30 frames in case it appears again. We set $\lambda=0.95$ in Eq.(\ref{our weight sum cost}). The IoU and appearance metric distance thresholds during association as 0.2 and 0.28, respectively. For results on MOT17 and MOT20, following\citep{bytetrack}, we use linear tracklet interpolation to compensate for in-perfections in the ground truth.
\subsection{Ablation Study}
In this part, we conducted ablation experiments on MOT17 with the main aims to analyze the effectiveness of our robust-of-tricks and to verify for each component what it contributes to the MOT. To avoid the influence on equity caused by the detector, we used ByteTrack's YOLOX-X MOT17 ablation study weights. We apply STP-DC, HG-FEN, CF-ECM and DA on baseline tracker. The results are presented in Table \ref{tab1}. The Baseline represents our re-implemented DeepSORT, without any guidance from addition modules.
\begin{table*}[!t]
	\begin{center}
		\caption{Ablation study on the MOT17 validation set for basic strategies, i.e., smooth trajectory predicts direction consistency(STP-DC), Hyper-grain feature extraction network(HG-FEN), Coarse-EMA Fine-EMA embedded cluster mechanism(CF-ECM), deep association trick(DA). All results obtained with the same parameters set. (best in bold).\\
			$^{\ast}$Ours reproduced results using TrackEval\cite{luiten2020trackeval}.}
		\label{tab1}
		
		\begin{tabular}{ c|cccc|cccc}
			\hline
			\textbf{Method}&\textbf{STP-DC} &\textbf{HG-FEN} &\textbf{CF-ECM} &\textbf{DA}&\textbf{MOTA($\%$)}$\uparrow$ & \textbf{HOTA($\%$)}$\uparrow$ & \textbf{IDF1($\%$)}$\uparrow$&\textbf{AssA($\%$)}$\uparrow$\\
			\hline
			\hline
			Baseline(DeepSORT$^{\ast}$)&-&-&-&-&64.00&56.54&66.06&58.16\\
			Baseline + column1&\checkmark&&&&76.50&64.03&75.52&62.77\\
			Baseline + column1-2&\checkmark&\checkmark&&&77.25&65.40&77.16&65.60\\
			Baseline + column1-3&\checkmark&\checkmark&\checkmark&&77.95&68.37&80.89&66.64\\
			Baseline + column1-4(Rt-Track)&\checkmark&\checkmark&\checkmark&\checkmark&\textbf{78.06}&\textbf{68.42}&\textbf{80.93}&\textbf{70.78}\\
			\hline
		\end{tabular}
	\end{center}
\end{table*}
\begin{table*}[!ht]
	\begin{center}
		\caption{comparison with preceding state-of-the-art methods on MOT17. The best results are highlighted in \textbf{\textcolor{red}{red bold}} and the second best results are highlighted with a \textcolor{blue}{\underline{blue underline}}}.
		\label{tab_mot17}
		
		\begin{tabular}{c|c|c|c|c|c|c|c|c|c|c|c}
			\hline
			
			\textbf{Method}&\textbf{MOTA}$\uparrow$&\textbf{IDF1}$\uparrow$&\textbf{HOTA}$\uparrow$&\textbf{MT}$\uparrow$&\textbf{ML}$\downarrow$&\textbf{FP}$\downarrow$&\textbf{FN}$\downarrow$&\textbf{Rcll}$\uparrow$&\textbf{AssA}$\uparrow$&\textbf{IDSW}$\downarrow$&\textbf{FPS}$\uparrow$ \\
			\hline
			\hline
			SORT\citep{sort}&43.1&39.8&34.0&12.5&42.3&28398&287582&49.0&31.8&4852&\textbf{\textcolor{red}{143.3}}\\
			TBooster\citep{TBooster}&61.5&63.3&50.5&26.4&32.0&-&-&-&52.0&2470&6.9\\
			%DAN\citep{DAN}&52.4&49.5&39.3&21.4&30.7&25424&234592&58.4&-&8431&6.3\\
			Tube-TK\citep{Tube-TK}&63&58.6&48.0&31.2&19.9&27060&177483&68.5&45.1&4137&3.0\\
			CTracker\citep{CTracker}&66.6&57.4&49.0&32.3&24.2&22284&160491&71.6&45.2&5529&6.8\\
			CenterTrack\citep{centertrack}&67.8&64.7&52.2&34.6&24.6&\textbf{\textcolor{red}{18498}}&160332&-&-&3039&3.8\\
			QuasiDense\citep{QuasiDense}&68.7&66.3&53.9&40.6&21.9&26589&146643&74.0&52.7&3378&20.3\\
			TraDeS\citep{trades}&69.1&63.9&52.7&36.4&21.5&\textcolor{blue}{\underline{20892}}&150060&73.4&50.8&3555&17.5\\
			MAT\citep{MAT}&69.5&63.1&53.8&43.8&18.8&30660&138741&75.4&51.4&2844&9.0\\
			%MOTR\citep{MOTR}&71.9&68.4&57.2&-&-&21100&136000&-&55.8&2115&-\\
			TransCenter\citep{TransCenter}&73.2&62.2&54.5&40.7&18.5&23112&123738&78.1&49.7&4614&1.0\\
			%GSDT\citep{GSDT}&73.2&66.5&55.2&41.6&17.4&26397&120666&78.6&51.0&3891&4.9\\
			FairMOT\citep{fairmot}&73.7&72.3&59.3&43.2&17.3&27507&117477&79.2&58.0&3303&25.9\\
			RelationTrack\citep{RelationTrack}&73.8&74.7&61.0&41.7&23.2&27999&118623&-&\textcolor{blue}{\underline{61.5}}&\textbf{\textcolor{red}{1374}}&8.5\\
			%PermaTrackPr\citep{PermaTrackPr}&73.8&68.9&55.5&43.8&17.2&28998&115104&79.6&53.1&3699&11.9\\
			CSTrack\citep{CSTrack}&74.9&72.6&59.3&41.5&17.4&23847&114303&79.7&57.9&3567&15.8\\
			GRTU\citep{GRTU}&74.9&75.0&\textcolor{blue}{\underline{62.0}}&\textcolor{blue}{\underline{49.7}}&18.8&32007&107616&80.9&\textbf{\textcolor{red}{62.1}}&\textcolor{blue}{\underline{1812}}&3.6\\
			CrowdTrack\citep{CrowdTrack}&75.6&73.6&60.3&46.5&\textbf{\textcolor{red}{12.2}}&25950&109101&80.7&59.3&2544&\textcolor{blue}{\underline{140.8}}\\
			%SiamMOT\citep{SiamMOT}&76.3&72.3&-&-&-&-&-&-&-&-&12.8\\
			CorrTracker\citep{corr-track}&76.5&73.6&60.7&47.6&\textcolor{blue}{\underline{12.7}}&29808&99510&-&-&3369&15.6\\
			TransMOT\citep{TransMOT}&76.7&\textcolor{blue}{\underline{75.1}}&61.7&-&-&36231&\textcolor{blue}{\underline{93150}}&-&59.9&2853&9.6\\
			ReMOT\citep{ReMOT}&\textcolor{blue}{\underline{77.0}}&72.0&59.7&\textbf{\textcolor{red}{51.7}}&13.8&33204&93612&\textcolor{blue}{\underline{83.4}}&57.1&2853&1.8\\
			%\hline
			%DeepSORT$^{*}$\citep{deepsort}&78.0&74.5&61.2&-&-&-&-&-&59.7&1821&13.8\\
			\textbf{Rt-Track(ours)}&\textbf{\textcolor{red}{79.5}}&\textbf{\textcolor{red}{76.0}}&\textbf{\textcolor{red}{62.1}}&\textbf{\textcolor{red}{51.7}}&17.2&27555&\textbf{\textcolor{red}{86034}}&\textbf{\textcolor{red}{84.8}}&60.6&2337&11.4\\
			\hline
		\end{tabular}
	\end{center}
\end{table*}

Based on the data in each row of Table \ref{tab1}, it can be observed that our STP-DC achieves significant improvements in several MOT metrics. This suggests that our trajectory consistency calculation strategy can effectively address tracking errors, leading to improved target position localization accuracy and long-term tracking and association correctness.

When we proceeded to apply the HG-FEN module, the tracking performance followed with improvements by 1.64\% IDF1 and 1.37\% HOTA. These improvements confirm that focusing on the hyper-grained features of the target benefit the enhanced distinguishability of the appearance features. It is also verified in Table \ref{tab4}. Meanwhile,it could be observed that CF-ECM also benefits the inference procedure. As shown, the performance of the tracker is improved by further 2.97\% HOTA and 3.73\% IDF1 compared to the metrics in the 3$_{rd}$ row, which indicates that our proposed Micro-embedding cluster mechanism plays a robust role. 

Finally the DA module also plays a robust matching role for our matching method SK-AS, which shows a significant improvement in the AssA metric, bringing a further improvement by 4.14\%. This result indicates that by connecting our proposed four robust tricks we can achieve a large gain in the tracker. 
\begin{table}[!h]
	\begin{center}
		\caption{The comparison of the hyper-grain feature extraction model with the baseline model can verify the validity of HG-FEN numerically. Tested on the MarKet1501 dataset.}
		\label{tab4}
		\begin{tabular}{c|ccc}
			\hline
			\textbf{Method}&\textbf{Rank@1}(\%)$\uparrow$&\textbf{mAP}(\%)$\uparrow$&\textbf{mINP}(\%)$\uparrow$\\
			\hline
			\hline
			Baseline(BoT$^{*}$)&94.3&86.3&61.2\\
			HG-FEN(ours)&94.8(\textcolor{red}{+0.5})&87.1(\textcolor{red}{+0.8})&62.5(\textcolor{red}{+1.3})\\
			\hline
		\end{tabular}
	\end{center}
\end{table}
\begin{table}[!h]
	\begin{center}
		\caption{Determine the number $k$ of linear fit points and distance $\Delta_t$ between $O$ and $A$ in the STP-DC using MOT17-val.}
		\label{tab3}
		\begin{tabular}{ c|  c  c  c  c }
			\hline
			& \textbf{HOTA}$\uparrow$ & \textbf{MOTA}$\uparrow$ & \textbf{IDF1}$\uparrow$ & \textbf{AssA}$\uparrow$\\
			\hline
			\hline
			$k=3$ , $\Delta_t=2$&67.891 &77.844 &80.189 &69.828\\
			\hline
			$k=5$ , $\Delta_t=2$&\textbf{68.046} &77.885 &\textbf{80.423} &\textbf{70.073}\\
			$k=5$ , $\Delta_t=3$&67.774 &77.942 &80.248 &69.474\\
			$k=5$ , $\Delta_t=4$&67.764 &77.790 &80.133 &69.444\\
			\hline
			$k=7$ , $\Delta_t=5$&67.716 &77.898 &80.037 &69.249\\
			$k=7$ , $\Delta_t=6$&67.698 &77.818 &79.929 &69.207\\
			\hline
			$k=9$ , $\Delta_t=7$&67.746 &\textbf{77.990} &80.253 &69.422\\
			$k=9$ , $\Delta_t=8$&67.812 &77.888 &80.039 &69.580\\
			\hline
		\end{tabular}
	\end{center}
\end{table}
\subsection{$k$ and $\Delta_t$ in STP-DC.}
For STP-DC, we consider that the linear fit of the previous period is meaningful for the smoothed trajectory of the later period. 

Therefore, we need to determine the number of base point. Besides, the time interval $\Delta_t$ between positions $O$ and $A$, in Eq.\ref{KF's correct meas}, has an important influence on calculating the directional consistency of the trajectory, as in Eq.\ref{Directional consistency calculation}, and reduces the impact of accumulated KF errors. As shown in the first element of the Table \ref{tab3}, enumerates the effect of the combination of number for fitting points  $k$ and time distance $\Delta_t$ on the MOT metrics. Based on the results in the table, better tracking performance is achieved when the number of base point is defined as 5 and the $O$ and $A$ time distance is 2.
\begin{table*}[!ht]
	\begin{center}
		\caption{comparison with preceding state-of-the-art methods on MOT20. The best results are highlighted in \textbf{\textcolor{red}{red bold}} and the second best results are highlighted with a \textcolor{blue}{\underline{blue underline}}}
		\label{tab_mot20}	
		\begin{tabular}{c|c|c|c|c|c|c|c|c|c|c|c}
			\hline
			\textbf{Method}&\textbf{MOTA}$\uparrow$&\textbf{IDF1}$\uparrow$&\textbf{HOTA}$\uparrow$&\textbf{MT}$\uparrow$&\textbf{ML}$\downarrow$&\textbf{FP}$\downarrow$&\textbf{FN}$\downarrow$&\textbf{Rcll}$\uparrow$&\textbf{AssA}$\uparrow$&\textbf{IDSW}$\downarrow$&\textbf{FPS}$\uparrow$ \\
			\hline
			\hline
			SORT\citep{sort}&42.7&45.1&36.1&16.8&26.3&27521&264694&48.8&35.9&4470&\textbf{57.3}\\
			Tracktor++\citep{Tracktor++}&52.6&52.7&42.1&29.4&26.7&\textbf{\textcolor{red}{6930}}&236680&54.3&42.0&1648&1.2\\
			TBooster\citep{TBooster}&54.6&53.4&42.5&32.8&25.5&-&-&-&41.4&1674&0.1\\
			%ArTISI\citep{ArTISI}&53.6&51.0&41.6&31.6&28.1&7765&230576&55.4&40.2&1531&1.0\\
			%LPC\citep{LPC}&56.3&62.5&49.0&34.1&25.1&11726&213056&58.8&52.4&1562&0.7\\
			%MPNTrack\citep{MAATrack}&57.6&59.1&46.8&38.2&22.5&16953&201384&61.1&47.3&1210&6.5\\
			TransCenter\citep{TransCenter}&58.5&49.6&43.5&48.6&14.9&64217&146019&71.8&37.0&4695&1.0\\
			%ApLift\citep{ApLift}&58.9&56.5&46.6&41.3&21.3&17739&192736&62.8&45.2&2241&0.4\\
			FairMOT\citep{fairmot}&61.8&62.3&54.6&68.9&\textbf{\textcolor{red}{7.6}}&103440&88901&82.9&54.7&5243&13.2\\
			TransTrack\citep{TransTrack}&65.0&59.4&48.5&50.1&13.5&27197&150197&-&-&3608&7.2\\
			CorrTracker\citep{corr-track}&65.2&69.1&-&66.4&\textcolor{blue}{\underline{8.9}}&45895&146347&-&-&4653&8.5\\
			CSTrack\citep{CSTrack}&66.6&68.6&54.0&50.4&15.5&25404&144358&72.1&54.0&3196&4.5\\
			%SiamMOT\citep{SiamMOT}&67.1&69.1&-&-&-&-&-&-&-&-&4.3\\
			GSDT\citep{GSDT}&67.1&67.5&53.6&53.2&13.2&31507&135395&73.8&52.7&3230&0.9\\
			RelationTrack\citep{RelationTrack}&67.2&70.5&56.5&62.2&\textcolor{blue}{\underline{8.9}}&61134&104597&-&56.4&4243&2.7\\
			%SOTMOT\citep{SOTMOT}&68.6&71.4&57.4&64.9&9.7&57064&101154&-&57.3&4209&8.5\\
			CrowdTrack\citep{CrowdTrack}&70.7&68.2&55.0&55.0&12.1&21928&126533&75.5&52.6&3198&8.5\\
			StrongSORT++\citep{strongsort}&73.8&77.0&\textcolor{blue}{\underline{62.6}}&62.1&14.9&\textcolor{blue}{\underline{16632}}&117920&77.2&\textbf{\textcolor{red}{64.0}}&\textbf{\textcolor{red}{770}}&1.4\\
			%MAATrack\citep{MAATrack}&73.9&71.2&57.3&59.7&12.3&24942&108744&-&55.1&1331&14.7\\
			OCSORT\citep{oc-sort}&75.7&76.3&62.4&65.5&12.9&19067&105894&79.5&62.5&\textcolor{blue}{\underline{942}}&\textcolor{blue}{\underline{18.7}}\\
			%ReMOT\citep{ReMOT}&77.4&73.1&61.2&68.2&9.9&28351&86659&83.3&58.7&1789&0.4\\
			ByteTrack\citep{bytetrack}&\textcolor{blue}{\underline{77.8}}&75.2&61.3&69.2&9.5&26249&\textcolor{blue}{\underline{87594}}&\textcolor{blue}{\underline{83.1}}&59.6&1223&17.5\\
			Bot-SORT\citep{botsort}&\textcolor{blue}{\underline{77.8}}&\textcolor{blue}{\underline{77.5}}&\textbf{\textcolor{red}{63.3}}&\textcolor{blue}{\underline{70.3}}&9.6&24638&88863&82.8&62.9&1313&2.4\\
			%\hline
			%DeepSORT$^{*}$\citep{deepsort}&71.8&69.6&57.1&-&-&-&-&-&55.5&1418&3.2\\
			\textbf{Rt-Track(ours)}&\textbf{\textcolor{red}{77.9}}&\textbf{\textcolor{red}{78.4}}&\textbf{\textcolor{red}{63.3}}&\textbf{\textcolor{red}{70.5}}&9.2&28390&\textbf{\textcolor{red}{84940}}&\textbf{\textcolor{red}{83.6}}&\textcolor{blue}{\underline{63.1}}&1196&0.5\\
			\hline
		\end{tabular}
	\end{center}
\end{table*}

\subsection{Effectiveness of STP-DC}
We compared the baseline tracker with its effect after applying STP-DC. The tracking results in the complete video sequence are shown in Fig. \ref{fig1}. The trajectory plot in Fig. \ref{fig1}(a) exhibits significant fluctuations. Our analysis suggests that this is primarily due to the measurement vector being susceptible to environmental factors and camera movement. The disturbed measurement vectors, during the KF update, led to the accumulation of system errors, resulting in significant oscillations in the trajectory.

From Fig. \ref{fig1}(b), we found that our smoothing strategy overcame the issue observed in Fig. \ref{fig1}(a), resulting in smoother trajectory plots. During the trajectory association phase, the smoothed trajectories are the key to the success of STP-DC. Fig. \ref{fig1}(c) demonstrates that the application of STP-DC is accurate in tracking each object and can recover the object's identity even in the presence of occlusion.Our tracking results video is available on the web link\footnote{\url{https://youtu.be/sPaP7EXoUJY}}.
\begin{figure}[!h]	
	\centering
	\includegraphics[width=3.5in]{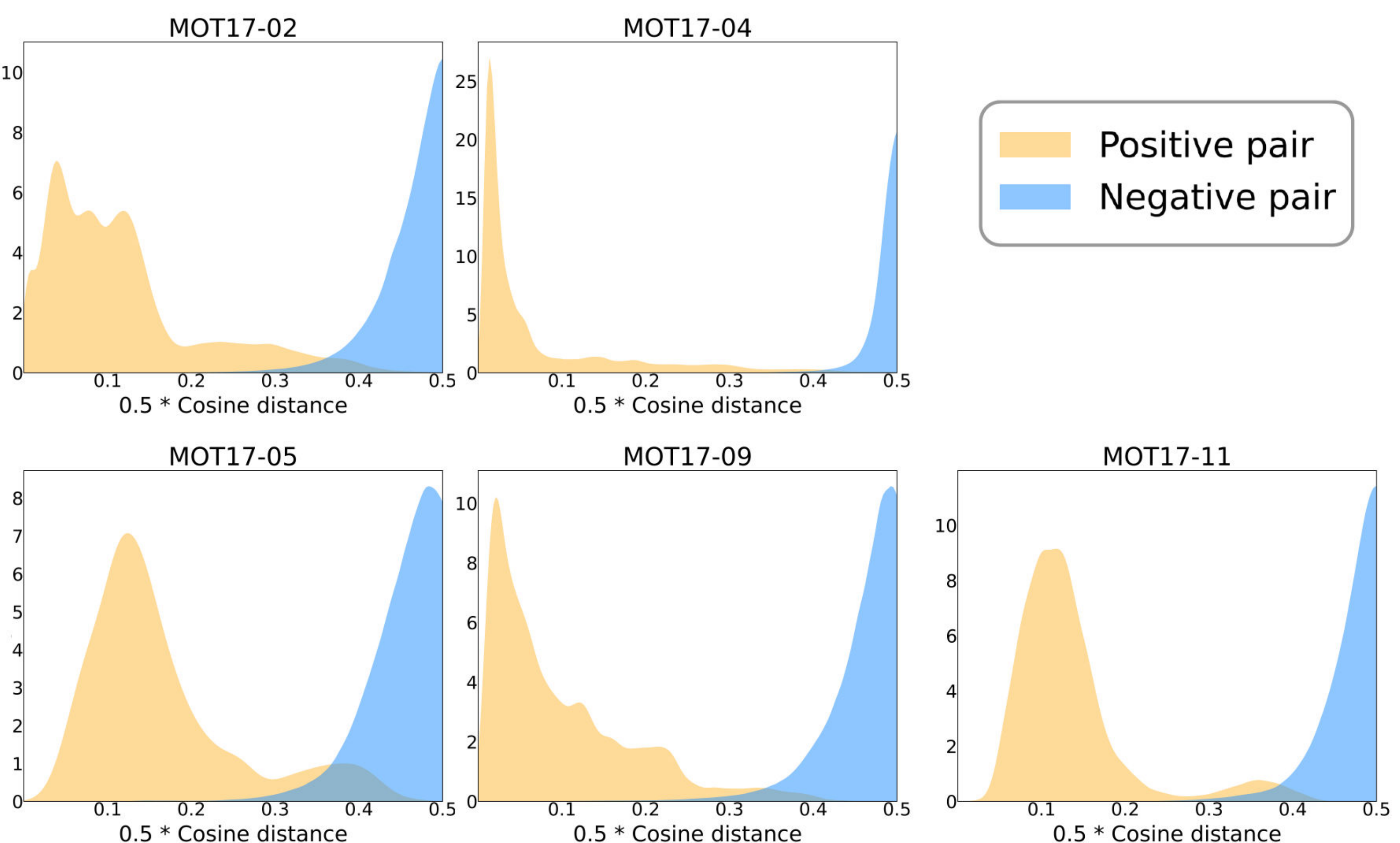}
	\caption{Distribution of affinity for intraclass and interclass appearance features using MOT17-Val.}
	\label{Distance Distribution Map}
\end{figure}
\begin{figure}[!h]	
	\centering
	\includegraphics[width=3.5in]{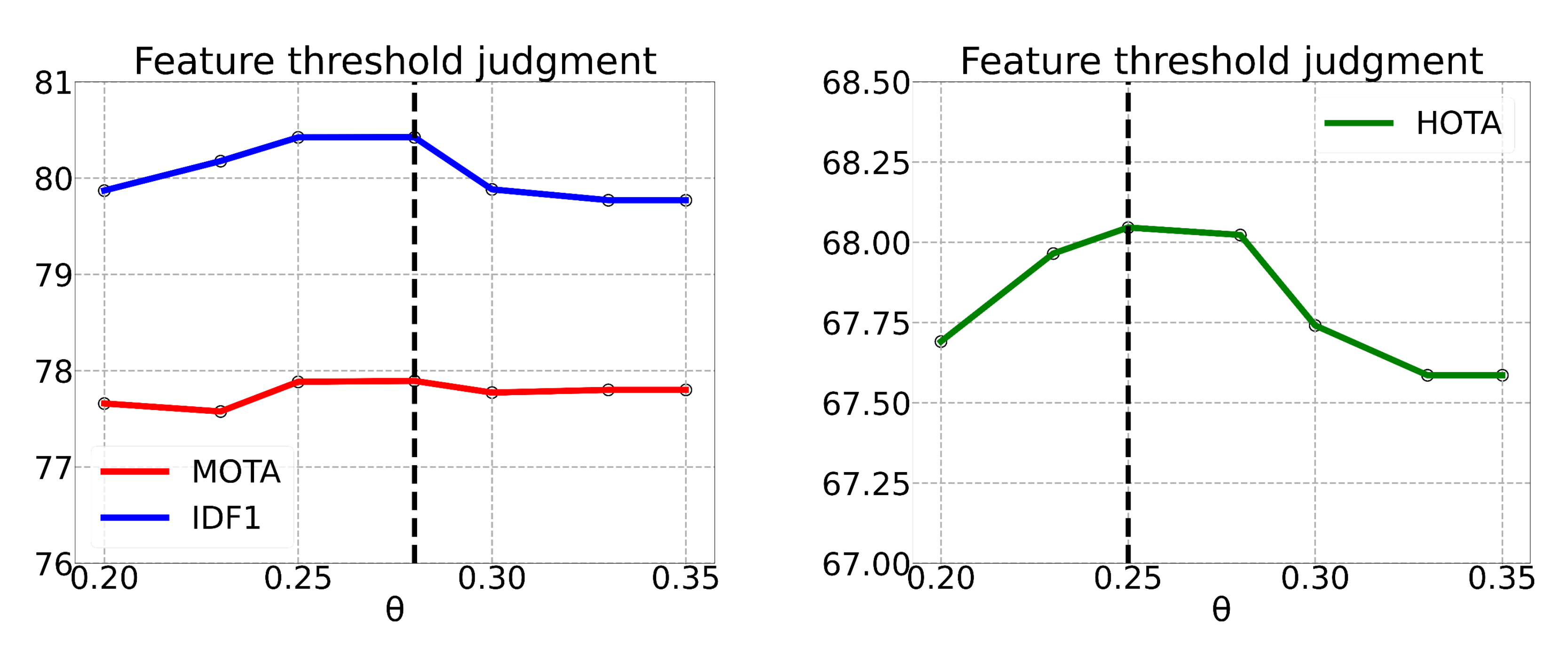}
	\caption{Determine Threshold $\theta$ of intra-class distance in HG-FEN using MOT17-Val.The virtual line represents the position of the best point.}
	\label{Feature thresholds}
\end{figure}
\subsection{Appearance affinity threshold $\theta$}
When using our HG-FEN to generate the cosine cost matrix, an affinity threshold is often required to reject unlikely associations. Fig.\ref{Distance Distribution Map} shows the intra-class distance distribution and inter-class distance tend to be between 0 to 0.2, 0.4 to 0.5, respectively. Meanwhile, Fig.\ref{Feature map} demonstrates that the pedestrian saliency regions generated by the proposed HG-FEN extract more pedestrian key features compared to other models, and this conclusion illustrates the robust discriminative power of our method. In order to obtain more accurate affinity thresholds $\theta$, we conducted experiments comparing the values of MOTA, IDF1 and HOTA between 0.2 to 0.4, as in Fig.\ref{Feature thresholds}, the position of the dashed guideline represents the best $\theta$ value of 0.28.
\begin{figure*}[!t]	
	\centering
	\includegraphics[width=7in]{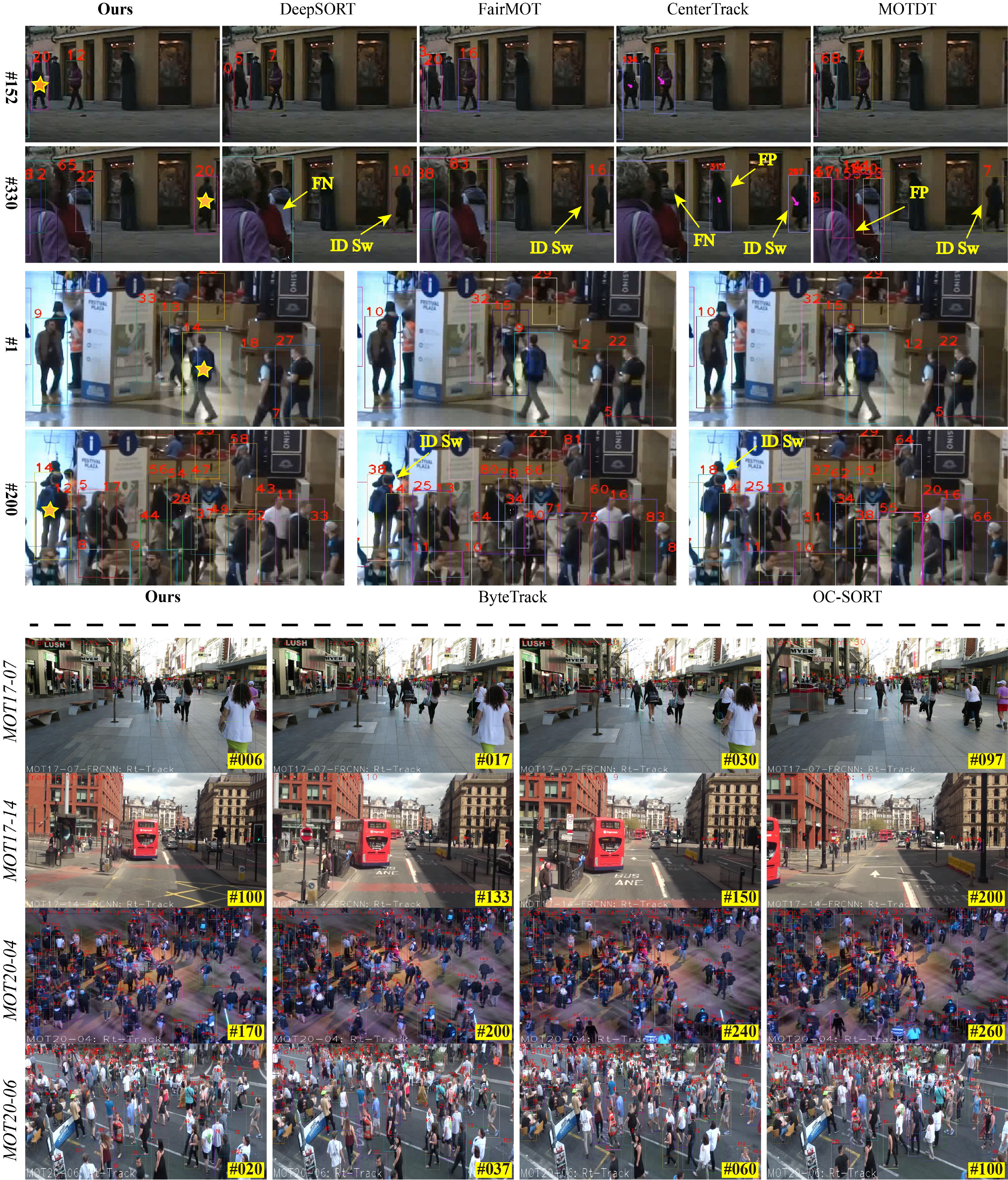}
	\caption{The top shows the tracking performance of our tracker compared to other trackers in complex scenarios, with video sequences taken from MOT17-02 and MOT20-01. \textcolor[rgb]{0.98039,0.68627,0.22745}{\ding{73}} represents the evaluation target. IDSW, FP and FN denote different kinds of false estimations which include identity switch, false negative and false positive, respectively. The bottom shows the qualitative results of our tracker on the MOT17 and MOT20 test sets.}
	\label{Robustness-analysis}
\end{figure*}
\begin{figure}[h]	
	\centering
	\includegraphics[width=3.5in]{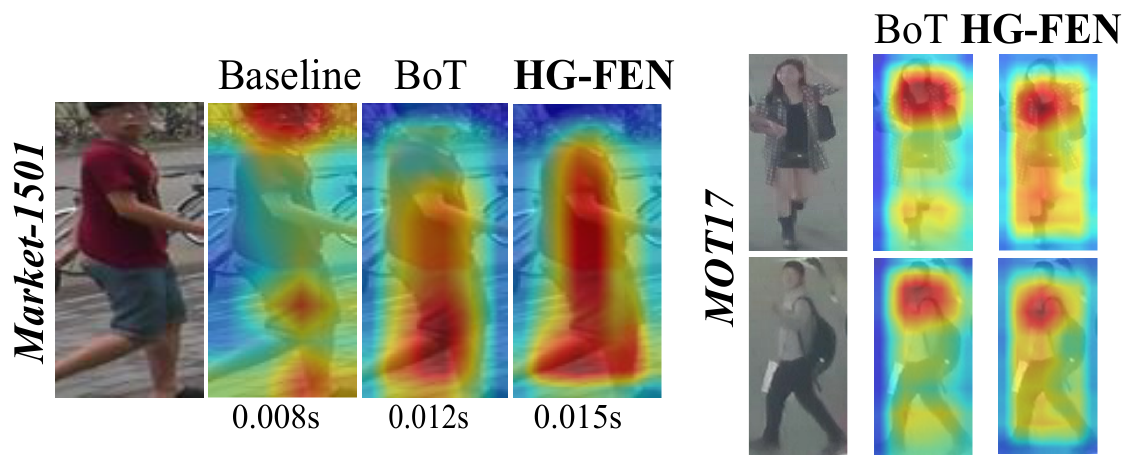}
	\caption{Visualizing object saliency regions using Market and MOT17 datasets. Baseline represents the original model in DeepSORT\citep{deepsort}. HG-FEN is the model we propose. Our model is able to focus on more target regions and only increases slightly in computation compared to BoT\citep{bot}.}
	\label{Feature map}
\end{figure}
\subsection{Robustness analysis in complex scenarios}

We visualized the performance of the previous tracker in complex scenarios, such as dense crowds and severe occlusions, shown at the top of Fig.\ref{Robustness-analysis}. The evaluation targets (with ID 20 and ID 14) reappeared after a long period of occlusion in both sequences. It can be observed that FairMOT and CenterTrack, using \citep{centernet} as a detector, exhibited poor performance in this situation. Both DeepSort and MOTDT\citep{MOTDT}, which use the same detector with us, also show varying degrees of errors. Relative to tracking in crowded situations, BateTrack and OC-SORT failed to maintain consistent target identity after a long period of occlusion (frame 1 to frame 200). Instead, Rt-Track can effectively identify the target and maintain its identity coherence. This demonstrates the robust performance of Rt-Track in complex scenarios. 

Fig.\ref{Robustness-analysis} bottom display the qualitative results of our Rt-Track on MOT17 and MOT20. These legends cover a variety of real-world complex scenarios, including camera motion, large and small samples, dense occlusion, and lighting changes, etc. Qualitative results show that Rt-Track is able to maintain its robustness and accurate differentiation of targets in these difficult scenarios.

\subsection{Main Results}
We evaluated the performance of the proposed tracker Rt-Track with several state-of-the art methods on benchmark datasets MOT17 and MOT20 test sets, following a private detection protocol, this result is reported in Table \ref{tab_mot17} and Table \ref{tab_mot20}, respectively. The official MOTChallenge challenge has always observed the principle of fairness, and thus all results are available from the official evaluation server. Notably, the speed of each method is dependent on the device with which they are implemented, and it is difficult to ensure absolute fairness of the FPS. For the two-stage tracker, the time consumption depends only on the tracking part.

\textbf{MOT17.} We compare the performance of trackers following the JBT\citep{centertrack,fairmot}, TBD\citep{sort,MAT,GRTU} and Transformer\citep{RelationTrack,TransCenter,TransMOT} tracking paradigms. According to the metric values in Table \ref{tab_mot17}, it can be seen that Rt-Track outperforms other previous artistic trackers in several major metrics. For instance, when compared to ReMOT, a tracker that uses appearance features to combine segmented trajectories, Rt-Track increases MOTA by 2.5\% (79.5\%-77.0\%), IDF1 by 4.0\% (46.0\%-42.0\%). In contrast to TransMOT, which treats target trajectories as sparse assignment graphs and uses a self-attentive mechanism to construct spatial relationships between targets, an increase of 2.8\% MOTA, 0.9\% IDF1 and a lower false negatives (FN) by 89034 is achieved. Compared to CorrTracker, which utilizes a local correlation module to model the topological relationship between targets and their context, there is a 4.1\% increase in HOTA, but our mostly lost targets (ML) are up by 4.5\%. Meanwhile, GRTU designs a multi-node tracking framework by designing virtual nodes to represent the missing detection under occlusion, compared to this method we achieve higher MOTA and IDF1, but the association accuracy (AssA) is reduced by 0.5\%. As expected, these results indicate the effectiveness of Rt-Track in potentially complex scenarios.

\textbf{MOT20.} Compared with MOT17, MOT20 is considered a difficult benchmark that includes more crowd scenes and occlusion situations, with an average of 170 pedestrians a frame. To demonstrate the robustness of our method in crowded and severely occluded situations, we compared it to a previously published state-of-the art tracker at MOT20. As shown in Table \ref{tab_mot20}, Rt-Track achieved the top-ranked performance on MOT20. Compared to the second-ranked Bot-sort\citep{botsort}, it has improved by 0.9\% (78.4\%-77.5\%) in IDF1 and has a lower IDSW, from 1313 to 1196. The increase is 2.0\% (63.3\%-61.3\%) in HOTA compared to the third ranked ByteTrack\citep{bytetrack}. Compared to OC-SORT\cite{oc-sort} and StrongSORT++\citep{strongsort}, they improved MOTA by 2.2\%, IDF1 by 2.1\% and MOTA by 4.1\% and IDF1 by 1.4\%, respectively, but IDSW was at a disadvantage. Comparatively, our method shows its state of the art performance in several metrics on MOT20, such as IDF1, MOTA, HOTA, MT, FN, Rcll, AssA, and IDSW, etc. This demonstrates the robustness of the proposed method in crowded and severely occluded situations for MOT tasks.

\section{Conclusion}
In this paper, we start with the difficulty of tracking in complex scenes faced by previous state-of-the-art methods. We propose four novel robust tricks, namely STP-DC, HG-FEN, CF-ECM, and SK-AS, which achieve higher tracking performance in occlusion and crowded scenarios. Among them, STP-DC implements a smooth trajectory and reduces tracking errors caused by cumulative KF system errors. HG-FEN extracts robust appearance descriptors, which can provide more discriminative information for association tasks in complex scenes. CF-ECM stores refined embedding symbols, which can recover lost trajectories after occlusion. The DA algorithm in SK-AS can bring more accurate association compared to the cascade algorithm in DeepSORT. 

By integrating STP-DC, HG-FEN, CF-ECM, SK-AS, and other techniques, our result tracker, named Rt-Track, achieves state-of-the-art performance on multiple object tracking benchmarks, i.e., MOT17 and MOT20.


\small
\bibliographystyle{IEEEtran}
\bibliography{references}

\newpage
%\begin{IEEEbiography}[{\includegraphics[width=1in,height=1.25in,clip,keepaspectratio]{fig2}}]{Yukuan Zhang}
% received a bachelor's degree in communication engineering from Polytechnic College of Hebei University of Science and Technology in 2021. He is pursuing the master's degree with the School of Physics and Electronic Information, Yunnan Normal University, Kunming, China.   His main research interests are deep learning, target detection, and target tracking.
%\end{IEEEbiography}

\vfill

\end{document}